\documentclass[letterpaper, 10 pt, journal, twoside]{ieeetran}

\usepackage{amssymb}  
\usepackage{dsfont}
\usepackage{graphicx}
\usepackage{amsmath} 
\usepackage{amsfonts}
\usepackage{multirow}
\usepackage[table,xcdraw]{xcolor}
\usepackage{booktabs}
\usepackage{url}
\usepackage{lipsum}
\usepackage{cite}
\usepackage{balance}

\DeclareMathOperator{\atantwo}{atan2}

\ifCLASSINFOpdf
\else
\fi

\begin{document}

\title{Sem-NaVAE: Semantically-Guided Outdoor Mapless Navigation via Generative Trajectory Priors}

\author{Gonzalo Olguín$^{1}$ and Javier Ruiz-del-Solar$^{1}$%
\thanks{This work was partially funded by FONDECYT project 1251823 and ANID project CIA250010.}%
\thanks{$^{1}$The authors are with the Department of Electrical Engineering \& the Advanced Mining Technology Center (AMTC), Universidad de Chile, Av. Tupper 2007, Santiago, Chile (e-mail: {\tt\footnotesize gonzalo.olguin@ing.uchile.cl}; {\tt\footnotesize jruizd@ing.uchile.cl}).}%
}


\maketitle
\begin{abstract}
This work presents a mapless navigation approach for outdoor applications. It combines the exploratory capacity of conditional variational autoencoders (CVAEs) to generate trajectories and the semantic segmentation capabilities of a lightweight visual language model (VLM) to select the trajectory to execute. Open-vocabulary segmentation is used to score and select the generated trajectories based on natural language, and a state-of-the-art local planner executes velocity commands. One of the key features of the proposed approach is its ability to generate a large variability of trajectories and select them to navigate in real-time. In real-world outdoor experiments, Sem-NaVAE achieves a 90\% success rate across routes of 120–240m in unseen environments, outperforming the nearest baseline by 10\% while remaining within 7\% of a map-based upper bound. A
video showing an experimental run of the system can be found
in \url{https://youtu.be/i3R5ey5O2yk}.
\end{abstract}


\IEEEpeerreviewmaketitle

\vspace{-10pt}
\section{Introduction}

\IEEEPARstart{A}{utonomous} navigation in mobile robotics remains one of the most enduring challenges in the field. Although traditional ``Map-Plan-Control'' approaches have proven effective in structured industrial settings, they rely on pre-defined metric maps and kinematic models that fail to adapt to rapid environmental changes or unforeseen obstacles.

To address these limitations, learning-based methods have gained popularity, offering superior performance in dynamic and unstructured environments by directly capturing complex behaviors from data \cite{surveyDLnav}. These approaches leverage deep learning to navigate changing terrains and variable lighting conditions, integrating visual and range sensors to enhance environmental perception \cite{endtoendnav, wvn}. 

However, the majority of direct supervised learning approaches face a critical limitation: \textit{unimodality}. In real-world navigation, multiple valid trajectories often exist to bypass an obstacle (e.g., passing a tree on the left or right). Traditional discriminative models tend to average these solutions, resulting in collisions or physically infeasible behaviors \cite{codevilla2018endtoend, florence2022implicit}. Here, generative models represent a paradigm shift. By learning the probability distribution of possible trajectories rather than a single deterministic output, these models capture the stochastic nature of real-world navigation, proposing multiple feasible and diverse motion hypotheses \cite{gupta2018social,sridhar2023nomad}.

However, despite their ability to ensure geometric feasibility, generative models often lack deep semantic understanding. A robot might generate valid paths through both a flowerbed and a paved path; while both are geometrically possible, the choice depends on social norms or contextual constraints that are difficult to encode in a traditional cost function. Recently, Vision-Language Models (VLMs) have emerged as powerful tools for equipping robots with common-sense reasoning \cite{song2025vlmsocial, shah2023lmnav}. While large-scale models bridge the gap between visual perception and complex logical decision-making, our approach leverages these models specifically for open-vocabulary semantic perception, enabling the interpretation of natural language constraints into actionable terrain preferences without the overhead of full logical reasoning.

\begin{figure}[t]
    \centering
    \includegraphics[width=\linewidth]{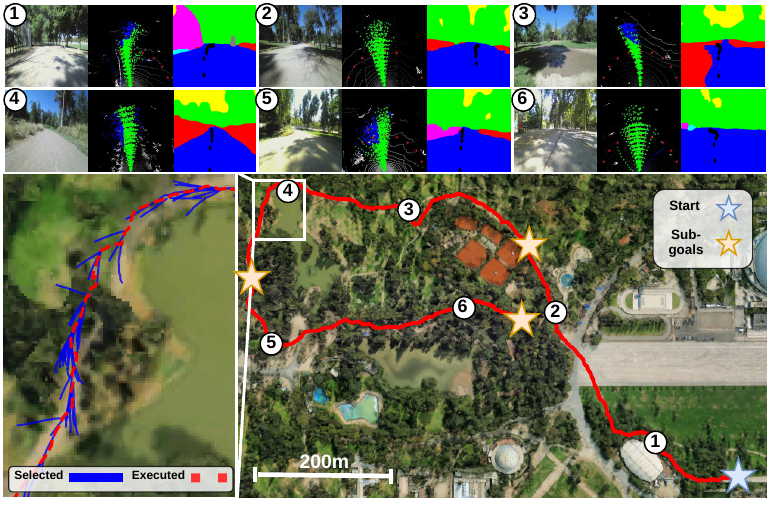}
     \vspace{-20pt}
    \caption{Examples of autonomous mapless navigation. Six timesteps are shown (1-6) with their corresponding selected and executed paths on a satellite image. Left rows show a First Person View (FPV) of the robot's onboard camera. Middle rows show an elevation map with the proposed trajectories (blue represents collisions and green dynamically feasible waypoints). Right rows show the selected trajectory projected onto an FPV costmap (in black).}
    \label{fig:trajectories}
    \vspace{-15pt}
\end{figure}

In this work, we propose a mapless global navigation framework that operates without a map during online inference. It combines the trajectory-generative capacity of Conditional Variational Auto-Encoders (CVAEs) in a \textbf{trajectory generation module} with the semantic segmentation capabilities of a lightweight VLM as a \textbf{trajectory selection module}. Unlike approaches dependent on maintenance-heavy global metric maps, our system plans long-horizon routes adaptively. We utilize a two-level architecture where our module acts as the high-level planner—using open-vocabulary segmentation to score trajectories and guide navigation based on natural language—while a state-of-the-art local planner executes velocity commands. An example of this framework, deployed in a wheeled UGV in a park environment, is illustrated in Fig.~\ref{fig:trajectories}.

The main contributions of this work are as follows:
\begin{itemize}
    \item \textbf{Perception-Constrained Generative Planning}: We introduce a CVAE architecture utilizing a custom Log-Mean-Exp (LME) multiple-choice training objective and PointNet-constrained decoding. This efficiently generates a  manifold ($>$200) of geometrically feasible trajectories that supply most of the available directions of movement.
    
    \item \textbf{Zero-Shot Semantic Trajectory Selection}: By strictly decoupling the geometric generator from an open-vocabulary segmentation model, our framework achieves zero-shot adaptability to novel user preferences and terrains, actively filtering kinematically infeasible routes.
    
    \item \textbf{Asynchronous Update Strategy}: We decouple low-frequency global planning from high-frequency local execution. The strategy introduces a formal hysteresis-based switching policy that prevents trajectory oscillation and resolves temporal occlusions by continuously re-evaluating the cost of the executing trajectory against newly generated candidates.
\end{itemize}

\begin{figure*}[t]
    \centering
    \includegraphics[width=\linewidth]{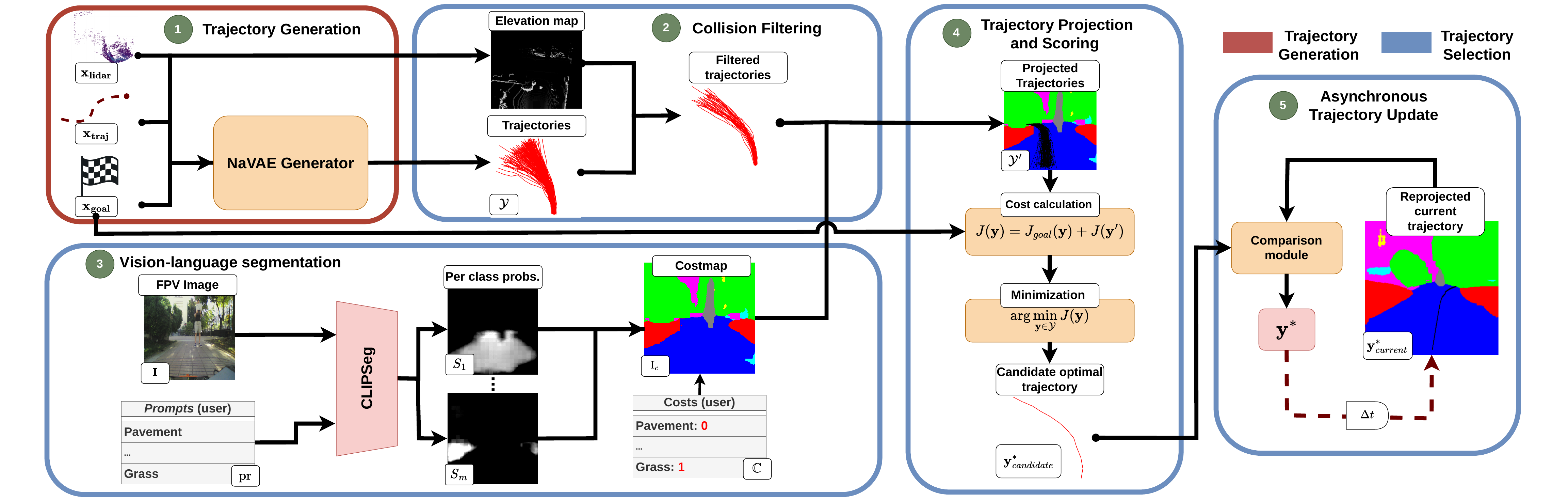}
    \vspace{-17pt}
    \caption{\textbf{Sem-NaVAE overview}: A generator module proposes a number of motion hypotheses based on sensory information. These hypotheses are then filtered by collisions and projected onto and FPV semantic map constructed with a lightweight VLM. Both semantic and goal-distance costs are used to then select the optimal trajectory. Instead of perdiodically updating to a new trajectory at every inference step, the new best one is compared with the currently executing one, only switching to the new when optimal.}
    \label{fig:system-overview}
    \vspace{-15pt}
\end{figure*}

\vspace{-10pt}
\section{Related Work}
\vspace{-1pt}

\subsection{Traversability Estimation and Semantic Perception}

To address physical interaction with semantics, early works like \cite{thrun2006stanley} correlated visual appearance with robot vibration. This has evolved into deep self-supervised learning, where proprioceptive signals (e.g. torque, velocity) automatically label visual data \cite{predictingterrain, travmapcontext} to later learn cost maps or navigation policies. Models trained from large-scale datasets, such as HDIF \cite{hdif2023} and recent work on interaction-aware navigation \cite{schoch2024insightinteractivenavigationsight} learn cost functions linking geometry to vehicle dynamics. The visual features of DINO \cite{dino} have been used in Velociraptor \cite{triest2024velociraptor} together with LiDAR to learn risk aware costmaps, and also in WVN \cite{wvn} to learn a traversability map online using only vision and velocity feedback. Approaches such as ViPlanner \cite{viplanner} fuse semantic labels with depth images to directly predict control actions and prefer safer terrains (e.g., asphalt instead of mud).

Although effective for stability and local safety, these traversability methods often struggle to generalize to semantically distinct environments not seen during training, as they lack a high-level conceptual understanding of the scene.

\vspace{-10pt}
\subsection{Generative Trajectory Models}

A common issue with learned planners in outdoor environments, particularly in complex scenarios, is the \textit{unimodality} of the proposed trajectories. To address this, generative models have emerged as the state of the art. Inspired by the inherent multimodality of human motion, numerous works have employed CVAEs \cite{trajectronplusplus, musevae} or diffusion models \cite{Jiang2023MotionDiffuserCM, Gu2022StochasticTP} to predict multiple motion hypotheses conditioned on observations such as pedestrian positions or local maps. Translating this paradigm to ego-centric robotic navigation, several studies leverage generative models to synthesize plausible trajectories directly from sensory data. In particular, MTG \cite{mtg} uses an attention-based CVAE to generate various predictions that maximize spatial coverage. This line of research is further advanced by DTG \cite{dtg}, which replaces CVAE with a diffusion-based predictor to improve the fidelity and stability of generated paths. Similarly, NoMaD \cite{sridhar2023nomad} employs a unified diffusion policy to model the joint distribution of subgoals and actions in image space, enabling robust exploration and navigation without global maps. 

However, even though these methods demonstrate superior capability in generating geometrically feasible and diverse motion plans, they typically lack a semantically grounded selection strategy, often relying on simple heuristics or geometric cost functions to choose the final trajectory.

\vspace{-10pt}
\subsection{Open vocabulary segmentation and VLMs for robotic navigation}

At a higher level, VLMs such as GPT-4 \cite{gpt4} have been integrated to interpret natural language commands \cite{song2025vlmsocial, Werby-hovsg}, allowing instruction-following behaviors (e.g., ``find a glass in the kitchen'') and even long-range navigation with behavioral cues \cite{chang2023goatthing, shah2023lmnav, vlmgronav}. However, such systems typically depend on the creation or existence of maps (whether topological or semantic), which presents a significant limitation in terms of scalability and deployment efficiency, as they necessitate prior exploration or the continuous maintenance of updated representations in dynamic environments.

Recent work applied VLMs to low-level control. Methods like PIVOT \cite{google2024pivot} and ConVOI \cite{sathyamoorthy2024convoi} use VLMs to select trajectories or modulate planner parameters based on context. Other works like BeHaV \cite{behav2025} use a combination of GPT with a lightweight VLM, CLIPSeg \cite{clipseg}, to perform open vocabulary segmentation, which enables faster inference for navigation. Closest to our work are TGS \cite{song2024tgs} and MOSU \cite{mosu}, which employ a ``Generate-and-Select'' paradigm: a CVAE generates geometrically feasible sub-goals, and a full-scale VLM selects the best candidate based on language alignment. MOSU adds a confidence metric and a semantic score based on closed vocabulary segmentation and QGIS routing for long range navigation. While these works establish a robust baseline, they rely on APIs, which have the problem of requiring an internet connection; consequently, they have an inference time on the order of seconds. 

\vspace{-5pt}
\section{Proposed Methodology}

An overview of the proposed methodology is shown in Fig. \ref{fig:system-overview}. We deliberately decouple geometric trajectory generation from semantic selection. This modularity provides zero-shot adaptability; changing navigation behaviors requires only a prompt update rather than retraining a joint model from scratch. Furthermore, by training the CVAE on diverse synthetic paths directed toward all reachable frontiers rather than single shortest-paths, the generator learns a manifold of geometrically feasible options, which are then filtered based on semantic segmentation.

\vspace{-10pt}
\subsection{Trajectory Generation}

\subsubsection{\textbf{Problem formulation}}

We consider a robotic agent navigating an unstructured, partially observable environment equipped with a multi-modal perception system. At time step $t$, the observations vector $\mathbf{x}$ is defined as $\mathbf{x} = \{ \mathbf{x}_{\text{lidar}}, \mathbf{x}_{\text{traj}}, \mathbf{x}_{\text{goal}} \},$ where:
\begin{itemize}
    \item $\mathbf{x}_{\text{lidar}} \in \mathbb{R}^{N_l \times N_p \times 4}$ represents the temporal sequence of the last $N_l$ processed point clouds. Each cloud consists of $N_p$ points defined by their spatial coordinates $(x, y, z)$ and intensity: $\{ \mathbf{s}^{\text{lidar}}_{t-N_l+1}, \dots, \mathbf{s}^{\text{lidar}}_{t} \}$.
    
    \item $\mathbf{x}_{\text{traj}} \in \mathbb{R}^{N_v \times 4}$ corresponds to the history of the last $N_v$ robot states, comprising position and linear velocity: $\{ \mathbf{s}^{\text{state}}_{t-N_v+1}, \dots, \mathbf{s}^{\text{state}}_{t} \}$.
    
    \item $\mathbf{x}_{\text{goal}} \in \mathbb{R}^{2}$ denotes the global navigation goal, expressed in polar coordinates $(\rho, \theta)$ relative to the robot's current reference frame.
\end{itemize}

The objective is to predict the future behavior of the agent over a prediction horizon $\tau$. We define the future trajectory as a sequence of spatial positions $\mathbf{y} = \{\mathbf{p}_{t+1}, \dots, \mathbf{p}_{t+\tau}\}$, where each $\mathbf{p}_k \in \mathbb{R}^2$ corresponds to 2D coordinates in the robot's base frame.

Due to inherent uncertainty in navigation and environmental interaction, the future is stochastic. Multiple physically plausible trajectories exist for a single historical context $\mathbf{x}$. Therefore, rather than seeking a single point estimate $\hat{\mathbf{y}} = f(\mathbf{x})$, we model future uncertainty via a conditional probability distribution $p(\mathbf{y} \mid \mathbf{x}).$

Since this underlying distribution can be highly complex and multimodal (encompassing distinct behaviors such as turning left or continuing straight), our goal is to learn a generative model parameterized by $\psi$ that approximates this distribution $p_\psi(\mathbf{y} \mid \mathbf{x})$.

Operationally, the model is required to generate a diverse set of $K$ hypothetical trajectories $\mathcal{Y} = \{\hat{\mathbf{y}}^{(k)}\}_{k=1}^{K}$ sampled from the learned distribution:
\begin{equation}
    \hat{\mathbf{y}}^{(k)} \sim p_\psi(\mathbf{y} \mid \mathbf{x}), \quad \text{for } k = 1, \dots, K.
\end{equation}

Thus, the problem reduces to maximizing the likelihood of the observed real trajectories under the model's predictive distribution, ensuring that the generated set $\mathcal{Y}$ covers the diverse modes of the true future distribution.

\begin{figure*}[t]
    \centering
    \includegraphics[width=\linewidth, height=4.6cm]{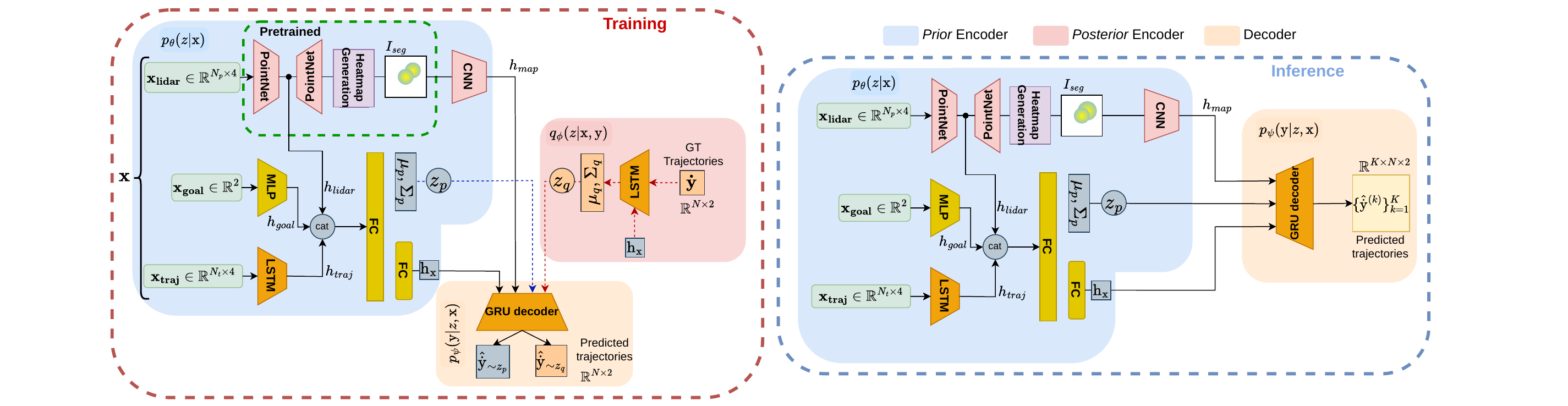}
     \vspace{-17pt}
    \caption{NaVAE architecture. The model takes as inputs consecutive pointclouds, past trajectory and navigation goal in polar coordinates. A pre-trained PointNet encoder serves as a feature extractor for the conditional value into the CVAE and as heatmap generator for teacher forcing in the GRU decoder. Taking $K$ samples from $p_\theta$ yields $K$ predicted trajectories.}
    \label{fig:arq_gen}
    \vspace{-15pt}
\end{figure*}

To achieve this, we propose an architecture based on CVAEs with a learned prior, following the framework proposed in \cite{trajectronplusplus} and \cite{musevae}. Fig. \ref{fig:arq_gen} illustrates the proposed NaVAE (Navigation VAE) architecture, highlighting the main modules: the pre-trained PointNet encoder, the prior encoder, the posterior encoder, and the stochastic decoder.

\subsubsection{\textbf{PointNet Pre-training \& Heatmap Generation}}

To extract environmental features and constrain the predicted trajectories using teacher forcing, the model integrates a pre-trained PointNet \cite{pointnet} that performs binary segmentation on the voxelized input point cloud $\mathbf{x}_{\text{lidar}}$, classifying points as \textit{traversable} or \textit{non-traversable}.

Ground truth labels are generated by projecting point coordinates onto a global semantic map (available only during training) filtering out points exceeding $1.5\times$ the robot's height (e.g., tree canopies). The network is trained via a weighted Cross-Entropy loss to address class imbalance.                                                                                                                     

To condition the trajectory decoder, the sparse point predictions are rasterized in a 2D grid with resolution $\delta = 0.2$m/px and a size of 18m$\times$18m. For each cell $(u, v)$, we compute the mean logit score to ensure invariance to the density of points. The final segmentation heat map $\mathbf{I}_{seg}$ is obtained by applying a Gaussian smoothing kernel $\mathcal{G}_{\sigma}$ ($\sigma=5$) to propagate semantic information to adjacent empty cells:
\begin{equation}
    \mathbf{I}_{seg} = \mathcal{G}_{\sigma} * (M^+ - M^- + \mathbf{1}),
\end{equation}
where $M^+$ and $M^-$ represent the occupancy grids for traversable and non-traversable logits, respectively.

\subsubsection{\textbf{NaVAE Network Architecture}}

Unlike standard CVAE formulations, our architecture introduces two fundamental innovations: a \textit{perception-constrained Decoder}—where a pre-trained PointNet heatmap explicitly restricts autoregressive velocity sampling—and a custom Log-Mean-Exp (LME) training objective. These operate alongside the standard \textit{Prior} and \textit{Posterior Encoders} to enable specialized, multiple-choice multimodal learning without suffering from mode-averaging.

The prior encoder $p_\theta(z|\mathbf{x})$ processes the context $\mathbf{x}$ to parametrize the latent prior $(\mu_p, \Sigma_p)$ (which then reparametrizes into $z_p$) and generates a context vector $\mathbf{h_x}$. It fuses the three input modalities (see  Fig. \ref{fig:arq_gen}):
\begin{itemize}
    \item \textbf{LiDAR:} The voxelized point cloud $\mathbf{x}_{\text{lidar}}$ is processed by the pre-trained PointNet backbone. It outputs a feature vector $h_{lidar}$ via a Fully Connected (FC) layer. Additionally, the segmentation logits are rasterized into a 2D Gaussian heatmap $\mathbf{I}_{seg}$, which is downsampled via CNN to obtain spatial features $h_{map}$ for the decoder.
    \item \textbf{Goal:} The polar goal vector $\mathbf{x}_{\text{goal}}$ is processed by a 2-layer MLP to produce $h_{goal}$.
    \item \textbf{State History:} Past trajectories $\mathbf{x}_{\text{traj}}$ (positions and velocities) are encoded by a single-cell LSTM to capture temporal dynamics and produce $h_{traj}$.
\end{itemize}
The features are concatenated as $\mathbf{h} = [h_{lidar}; h_{goal}; h_{traj}]$ and passed through LayerNorm and FC layers to produce the final context representation $\mathbf{h_x}$ and the latent parameters $(\mu_p, \Sigma_p)$.

The posterior encoder $q_\phi(z|\mathbf{x}, \mathbf{y})$, available only during training, encodes the ground truth future trajectory $\mathbf{y}$. We utilize a Bidirectional LSTM to process $\mathbf{y}$, allowing the latent state $z_q$ to capture global temporal dependencies (e.g., how the final destination influences initial maneuvering). The output is projected to the latent space parameters $(\mu_q, \Sigma_q)$.

The decoder $p_{\psi}(\mathbf{y}|z, \mathbf{x})$ operates as an autoregressive generator in velocity space to ensure kinematic smoothness. It is implemented as a GRU, as are many other works \cite{mtg, musevae}.

At each step $t$, the GRU receives the latent sample $z$ (both from $p_\theta$ and $q_\phi$ during training, just from $p_\theta$ during inference), the context $\mathbf{h_x}$, and the features of the spatial map $h_{map}$. It outputs the parameters of a Gaussian distribution over velocities $(\mu_{\dot{y}}, \Sigma_{\dot{y}})$. 
During inference, we sample velocities $\mathbf{v}_t$ from this distribution, filtering candidates that violate acceleration limits ($>0.5$ m/s$^2$) or fall into non-traversable regions of $\mathbf{I}_{seg}$ after projection. The final trajectory positions are obtained via numerical integration: $\mathbf{p}_{t+1} = \mathbf{p}_t + \mathbf{v}_t \Delta t$.

\subsubsection{\textbf{Training Objective}}

The training objective is based on a CVAE formulation with a learned prior. We minimize a modified Evidence Lower Bound (ELBO) that includes the Kullback-Leibler (KL) divergence between the prior $p_\theta$ and posterior $q_\phi$, along with the reconstruction likelihoods. Following \cite{musevae}, we enforce reconstruction from the posterior samples $z_q$ and the prior samples $z_p$ to align both distributions with the ground truth:
\begin{equation}
    \resizebox{\linewidth}{!}{
    $\mathcal{L}_{\text{ELBO}} = -\mathbb{E}_{z_p}[\log p_\psi(\mathbf{y}|z_p, \mathbf{x})] -\mathbb{E}_{z_q}[\log p_\psi(\mathbf{y}|z_q, \mathbf{x})] + \beta D_\mathrm{KL}(p_\theta || q_\phi). 
    $}\label{eqn:elbo}
\end{equation}

However, standard likelihood maximization tends to average conflicting modes, leading to valid but unrealistic ``mean'' trajectories. But we are interested in the general case where the dataset contains $M$ valid ground truth trajectories $\{\mathbf{y}^{(m)}\}_{m=1}^{M}$ for similar contexts. To capture this diversity, we reformulate the reconstruction loss for the \textit{prior} using a \textit{log-mean-exp} strategy.
This is mathematically analogous to Importance Weighted Autoencoders (IWAE) \cite{iwae}, where moving the summation inside the logarithm provides a tighter variational lower bound. Operationally, this acts as a \textit{soft Winner-Takes-All} mechanism: the gradient is dominated by the hypothesis $\mathbf{y}^{(m)}$ that best matches the ground truth. This allows the model to specialize in specific modes without being penalized for not predicting mutually exclusive alternatives.

Crucially, this formulation applies only to the prior (which must cover all modes). The posterior $q_\phi$, which is conditioned on a specific ground truth during training, uses standard likelihood. The final reconstruction loss is:
\begin{multline}
    \mathcal{L}_{\text{recon}} = -\log \left( \frac{1}{M}\sum_{m=1}^{M} p_\psi(\mathbf{y}^{(m)}|z_p, \mathbf{x})\right)\\ -\mathbb{E}_{z_q}[\log p_\psi(\mathbf{y}|z_q, \mathbf{x})].
    \label{eq:navae-recon-loss}
\end{multline}

Furthermore, to explicitly discourage non-navigable predictions, we introduce a collision loss $\mathcal{L}_{\text{col}}$ similar to the one proposed in \cite{mtg}. Trajectories are projected onto the local semantic ground truth map $\mathbf{M}_{sem}$. We apply a Gaussian Kernel $\mathbf{GK}(\cdot)$ over the map to smooth the gradients, penalizing waypoints that fall into non-traversable regions:
\begin{equation}
    \mathcal{L}_{\text{col}} = \log\left(\sum_{(p_x, p_y)\in\mathbf{y}'} \mathbf{GK}(\mathbf{M}_{\text{sem}}(p_x, p_y))\right).
    \label{eqn:col-loss}
\end{equation}

The total objective function is a weighted sum: $\mathcal{L}_{\text{Total}} = \beta D_\mathrm{KL}(p_\theta || q_\phi) + \mathcal{L}_{\text{recon}} + \lambda \mathcal{L}_{\text{col}},$ where $\beta$ and $\lambda$ are regularization hyperparameters.

\vspace{-10pt}
\subsection{Trajectory Selection}

\subsubsection{\textbf{Collision filtering}} 

Since the generated trajectories can have kinematic failures, a post-generation filter is required to ensure feasibility. For this, we construct an instantaneous local elevation map from LiDAR data, covering a forward-facing area of 18m $\times$ 18m with a resolution of $0.1$m/px.

A trajectory is classified as a collision if the pitch angle between any two consecutive waypoints exceeds a safety threshold $\theta_{max}$. For a waypoint $w_i$ and its predecessor $w_{i-1}$, the inclination is calculated as: 
\begin{equation}
    \theta_i = \atantwo\left(z_i - z_{i-1}, \sqrt{\Delta x^2 + \Delta y^2}\right),
    \label{eq:filtro-col}
\end{equation}

where $\Delta x, \Delta y$ are the planar displacements, and $z_i$ is determined by the maximum elevation value within the robot's square footprint $d_{foot}$ centered at $(x_i, y_i)$. Any trajectory containing a segment where $\theta_i > \theta_{max}$ is discarded. This slope-based constraint effectively filters out step obstacles and untraversable gradients while accounting for the robot's physical dimensions.

\subsubsection{\textbf{Vision-language segmentation}}
Following the generation of dynamically feasible trajectories, the subsequent task is to identify the candidate that best aligns with user-imposed constraints. To achieve this, we employ CLIPSeg \cite{clipseg}, a lightweight vision-language segmentation model. Specifically, we utilize the off-the-shelf, pre-trained base model ($\sim$130M parameters) without any task-specific fine-tuning.

Given a First-Person View (FPV) input image $\mathbf{I}$ and a set of open-vocabulary textual constraints $\mathbf{pr} = \{pr_1, pr_2, \ldots, pr_m\}$, CLIPSeg yields a corresponding set of dense, continuous probability maps $\mathbf{S} = \{S_1, S_2, \ldots, S_m\}$. To query the model, we utilize a minimalist prompt design, avoiding complex prompt engineering. Every constraint $pr_i$ is formatted using a strict template: \textit{``a photo of a [class]''}. 

Here, each map $S_i$ represents the pixel-wise probability (normalized via a sigmoid function) of the prompted class being present within $\mathbf{I}$. These classes typically correspond to semantic features relevant to unstructured outdoor environments, such as ``grass,'' ``person,'' etc.. Crucially, these constraints are modular and can be dynamically swapped depending on the specific navigation task.

To synthesize a unified semantic cost map $\mathbf{I}_c$, we assign a user-defined scalar cost coefficient to each textual class: $\mathbb{C} = \{c_1, c_2, \ldots, c_m\}$. Our post-processing pipeline is intentionally lightweight: we apply a pixel-wise $\arg\max$ across all probability maps to determine the dominant class at each pixel, which is then mapped to its corresponding cost:
\begin{equation}
    \mathbf{I}_c(u, v) = \mathbb{C}\left[ \arg\max_{i} \left( S_i(u, v) \right) \right],
    \label{eq:vlm_costmap}
\end{equation}

where $(u, v)$ denotes the pixel coordinates and $i \in \{1, \dots, m\}$. The resulting matrix $\mathbf{I}_c$ provides a dense semantic representation of the environment's traversability costs directly aligned with the user's intent.

\subsubsection{\textbf{Trajectory Projection and Scoring}}

The generated trajectories $\mathcal{Y}$, initially defined in the robot's local frame, must be evaluated against the semantic cost map $\mathbf{I}_c$. We explicitly opt for a 3D-to-2D trajectory projection approach for two critical reasons. First, projecting dense trajectories onto the dense 2D image plane prevents the severe loss of semantic information that occurs when mapping to sparse LiDAR point clouds, which suffer from inherent vertical gaps between scan lines. Second, direct projection relies on simple matrix multiplication, maintaining the low computational latency required for real-time execution. First, a trajectory $\mathbf{y}$ is transformed into the camera frame using the extrinsics $\mathbf{T_{r-c}}$ and then projected onto the image plane using the camera intrinsics: $\mathbf{y}' = K [I_3 | 0] \mathbf{T_{r-c}} \mathbf{y}.$ Here, $\mathbf{y}'$ contains the pixel coordinates $(p_x, p_y)$ corresponding to the trajectory waypoints.

We define the semantic cost $J(\mathbf{y}')$ as the discounted sum of costs, sampled from $\mathbf{I}_c$. A critical challenge in this 3D-to-2D projection is the loss of depth information; a valid path behind an obstacle might project onto the obstacle's pixels (occlusion). To handle this, we employ a masking mechanism:
\begin{equation}
    J(\mathbf{y}') = \sum_{j=1}^{N} \gamma^j \left[ \mathbf{I}_c(p_x^j, p_y^j) \cdot \mathds{1}_{u_j=0} + C_u \cdot \mathds{1}_{u_j=1} \right],
\end{equation}

where $\gamma \in [0, 1]$ is a discount factor for future waypoints, and $C_u$ is a fixed penalty. The binary mask $u_j$ indicates an occlusion; it is set to $1$ if the semantic cost at the pixel exceeds a threshold $T_{occ}$ (distinguishing between "soft" traversable terrain and strict obstacles, see Section \ref{sec:metrics}), and $0$ otherwise.

Finally, to ensure progress toward the target, we incorporate a geometric goal term $J_{goal}$:
\begin{equation}
    J_{goal}(\mathbf{y}) = \alpha_1 \log (1+d(\mathbf{y}, \text{goal})) + \alpha_2 |\theta_{\text{end}}|\pi^{-1},
\end{equation}

where $d(\cdot)$ is the Euclidean distance to the goal and $\theta_{\text{end}}$ represents the heading alignment error of the last waypoint. The optimal trajectory $\mathbf{y}^*$ is selected by minimizing the~joint~cost:
\begin{equation}
    \mathbf{y}^* = \arg\min_{\mathbf{y}\in\mathcal{Y}} \left( J(\mathbf{y}') + J_{goal}(\mathbf{y}) \right).
    \label{eq:selection-cost-function}
\end{equation}

\subsubsection{\textbf{Asynchronous Trajectory Update}}\label{sec:traj-update}

Standard planners often employ a receding horizon strategy, executing only the immediate next step. However, given the semantic uncertainty of our 3D-to-2D projection, blindly discarding the long-term plan is suboptimal. We propose an asynchronous update scheme that balances stability with reactivity.

The system operates on two decoupled frequencies: geometric generation ($f_{gen}$) and semantic perception ($f_{clip}$). While the robot executes a selected trajectory ($\mathbf{y}_{curr}$), its cost is not static; it is re-calculated at ($f_{clip}$) based on the latest CLIPSeg segmentation. Specifically, the costs of already traversed waypoints are frozen, while future waypoints are re-projected onto the new cost map. This allows the system to refine its estimate as the robot approaches a target, resolving initial occlusions or correcting misclassifications (e.g., distinguishing a shadow from an obstacle) without requiring geometric regeneration.

To determine when to switch trajectories, we compare the re-evaluated cost of $\mathbf{y}_{curr}$ against the best candidate from a newly generated set $\mathcal{Y}$. To prevent oscillations, a switch is triggered only if the new candidate improves the cost by a hysteresis factor $\epsilon > 0$: $J(\mathbf{y}_{new}) < J(\mathbf{y}_{curr}) - \epsilon.$

This hybrid design ensures that navigation decisions are always based on the most recent visual evidence, while minimizing computational overhead.

\vspace{-10pt}
\section{Results}

\subsection{Implementation and Experimental Setup}

To train the generative model, we collected a real-world dataset using a Husarion Panther UGV equipped with a 64-channel Ouster LiDAR, a generic RGB camera, IMU, GPS-RTK and a NVIDIA RTX4060 laptop. Data collection was carried out on the engineering campus of the University of Chile, an unstructured urban environment characterized by pedestrian paths, vegetation, and significant elevation changes.

To enable multimodal learning, we implemented a post-processing pipeline using the recorded ROS bags. For each data sample, we utilized the ROS NavFN planner to generate synthetic valid paths from the robot's current pose to all reachable frontiers within a radius $R_{max}=15$m on a static semantic map, constructed with AMCL and satellite image overlays. Candidate paths were filtered to ensure diversity, retaining only those with a Final Displacement Error (FDE) $> 0.5$m relative to others (averaging 5 paths per sample).

The final dataset comprises 1,706 samples ($\sim$ 8,500 candidate trajectories) derived from 15 full teleoperation runs sliced at 0.5s intervals. Each sample contains a slice of the global map, the ground truth trajectories, voxelized LiDAR and corresponding odometry data aligned with the map slice.

It is important to note that the semantic maps, AMCL and NavFN are utilized \textbf{exclusively} for training supervision and offline evaluation; they are \textbf{not available} to the robot during online inference.

We choose $N_l=$3, $N_p=$2,560 and use the last 5 seconds of trajectory data at 0.5 second intervals, with velocity normalized by the maximum of 2m/s. We predict $k=200$ trajectories in each step with $N_w=12$ waypoints each at a temporal distance of 1 second. The feature size is set to 256 and the latent space dimension $N_z$ to 512.

We train 1,000 epochs on an NVIDIA RTX4070 device for about 7 hours. We use AdamW optimizer with a learning rate of 0.001 and an exponential scheduler every 10 epochs. We use $\beta$ as a KL annealing factor to regularize the latent space after a good reconstruction performance is achieved. The collision loss weight $\lambda$ is set to 10.

\vspace{-10pt}
\subsection{Quantitative results}\label{sec:metrics}

To assess the efficiency and effectiveness of the navigation system, we employ the following metrics:

\begin{itemize}
    \item \textbf{Success Rate (SR):} The percentage of episodes in which the robot reaches the goal within a 5 meter threshold ($\sim$~2-4\% error for 120-240 m tests).
    
    \item \textbf{Fréchet Distance (FD) \cite{frechet}:} We measure similarity via FD between a human teleoperated trajectory and fully autonomous run in the same route.
    
    \item \textbf{Executed Path Traversability (EPT):} The percentage of the robot's \textit{executed} trajectory that remains within preferred ``soft'' traversable zones. We distinguish between \textit{strict} obstacles (walls, trees) that cause collision/failure and \textit{soft} obstacles (grass, dirt) that are traversable but may be semantically undesirable 

    \item \textbf{Non-Traversable Rate (NTR):} The percentage of \textit{generated} waypoints that fall into non-traversable regions (both soft and strict), as defined in \cite{mtg}. This evaluates the model ability to learn semantic constraints during inference, independent of the final selected path.

    \item \textbf{Navigation Time Ratio ($\mathbf{T_{ratio}}$):} The ratio between the autonomous navigation time and the teleoperated time.

    \item \textbf{Recovery Behaviors (\#RB):} A turn-in-place recovery is added when no feasible trajectories are found. \#RB measures the average triggers per run (rounded). 
\end{itemize}
To evaluate out-of-distribution (OOD) generalization, the 5 testing configurations (ranging from 120 to 240 meters with different ground types and vegetation percentages) were conducted in a strictly unseen sector of the engineering campus. Quantitatively, this unseen sector differs from the training distribution by containing 41\% fewer pedestrian paths, 30\% higher general obstacle density, and 60\% more vegetation, relatively. Furthermore, the semantic distribution introduces novel terrain features; pedestrian paths shift from homogeneous tiles in training to a mix of tiles, pavement, and gravel during evaluation. Each trial was repeated 6 times.

The baseline selection parameters $\gamma, \alpha_1, \alpha_2$ are set to 0.8, 2, 0.2, respectively. CLIPSeg is deployed locally, where specific semantic classes ([\textit{class}]) inserted into the prompt template, and their assigned costs, are: '\textit{pavement}' (0), '\textit{tree}' (3), '\textit{grass}' (2), '\textit{wall}' (3), '\textit{stairs}' (3), '\textit{person}' (3), '\textit{hole}' (3), '\textit{sky}' (4), and $C_u$, $T_{occ}$ are both set to 2, as this is the maximum soft-traversable class value.

The update frequencies $f_{gen}, f_{clip}$ are set to 0.5~Hz and 2.5~Hz, respectively, while the local planner \cite{leiva2024combining} handles dynamic obstacle avoidance at 50~Hz. Table \ref{tab:inference-times} details the inference times for each module. As discussed in Section \ref{sec:traj-update}, these frequencies are decoupled; thus, the system's overall semantic update rate is bounded by $f_{clip}$.

\begin{table}[ht]
\vspace{-5pt}
\caption{Inference times of each major module for 200 predictions.}
\label{tab:inference-times}
\resizebox{\linewidth}{!}{%
\begin{tabular}{@{}ccccc@{}}
\toprule
\textbf{Obs. processing [ms]} & \textbf{NaVAE [ms]} & \textbf{CLIPSeg [ms]} & \textbf{Col. filter [ms]} & \textbf{Costmap \& costs [ms]} \\ \midrule
$10\pm 0.74$                       & 103$\pm 8$            & 398$\pm13$              & 0.28$\pm 0.89$                 & 3$\pm0.12$                      \\ \bottomrule
\end{tabular}%
}
\vspace{-5pt}
\end{table}

We evaluate our framework against a comprehensive set of baselines spanning diverse navigation paradigms. \textbf{Nav2} is a map-based method that establishes a geometric performance upper bound. For modern diffusion policies, we compare against \textbf{NoMaD} (topological map). We use \textbf{MTG} as the primary CVAE baseline, alongside \textbf{MTG'}—a variant retrained with our PointNet module. We also compare to our own implementation of \textbf{VL-TGS} \cite{song2024tgs}, limiting the output number to the 6 (non-colliding) most diverse trajectories in~terms~of~FDE.

\begin{figure*}[t!]
    \centering
    \includegraphics[width=\linewidth]{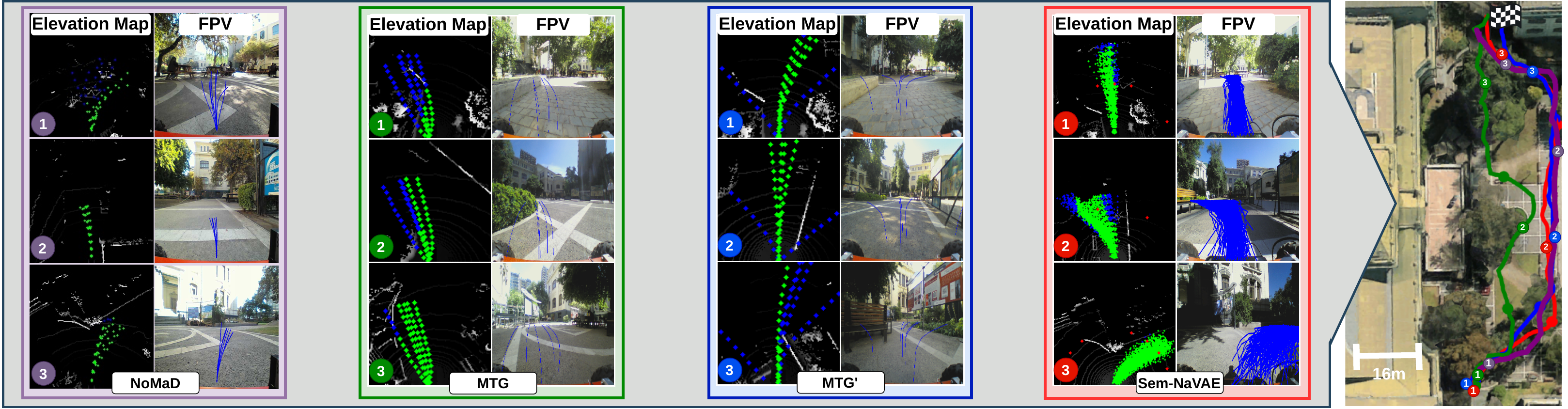}
     \vspace{-15pt}
    \caption{Comparison of baselines NoMaD \cite{sridhar2023nomad}, MTG \cite{mtg}, its PointNet modification MTG', and our system Sem-NaVAE. Left rows show an elevation map with the proposed trajectories (blue represents collisions and green free-space) and right rows show the generated trajectories projected onto an FPV image. Trials with the best results are plotted.}
    \label{fig:generated-trajectories}
    \vspace{-15pt}
\end{figure*}

As shown in Table \ref{tab:results}, Sem-NaVAE outperforms the metric-mapless baselines in SR, EPT, and $\mathbf{T_{ratio}}$, achieving improvements of 10\%, 0.2\%, and 44\%, respectively, while performing competitively with Nav2. The comparable NTR between Sem-NaVAE and MTG' reflects a deliberate trade-off: the CVAE is trained to generate a large volume of diverse trajectories, including some that explore non-traversable regions. The selection module then filters these, resulting in higher EPT scores than MTG' and NoMaD. As observed in Fig. \ref{fig:generated-trajectories}, Sem-NaVAE not only generates a larger volume of trajectories but also produces more complex curvatures. This enables a smoother global path and reduces reliance on recovery behaviors. Regarding baseline failure analysis, MTG exhibits low output variability due to the LiDAR encoder's sensitivity to training domain shifts. While the pre-trained PointNet in MTG' improves this—achieving the best NTR (22.1)—it still lacks the geometric curvature required to navigate tight spaces. 

\begin{table}[bt]
\vspace{-10pt}
\caption{Comparison of baselines and our model, Sem-NaVAE. }
\label{tab:results}
\resizebox{\linewidth}{!}{%
\begin{tabular}{@{}ccccccc@{}}
\toprule
\textbf{Model} & \textbf{SR $\uparrow$} & \textbf{FD $\downarrow$} & \textbf{EPT\% $\uparrow$} & \textbf{NTR\% $\downarrow$} & \textbf{$\mathbf{T_{ratio}}$ $\downarrow$} & \textbf{\#RB $\downarrow$} \\ \midrule
Nav2           & 0.97$\pm$0.17           & 5.85$\pm$0.82           & 98.6$\pm$0.93          & -               & 1.28$\pm$0.29           & -           \\ \midrule
NoMaD          & 0.80 $\pm$0.40          & \textbf{5.19$\pm$0.98}        & 94.8$\pm$0.61           & 22.9$\pm$1.98            & 3.26$\pm$0.70           & \textbf{2$\pm$1}           \\
MTG'           & 0.63$\pm$0.48           & 6.83$\pm$0.87           & 90.7$\pm$6.81          & \textbf{22.1$\pm$2.32}   & 2.27$\pm$0.36           & 3$\pm$1           \\
MTG            & 0.17$\pm$0.37           & 12.8$\pm$1.21           & 77.5$\pm$13.4          & 32.7$\pm$7.67            & 2.62$\pm$0.56           & 4$\pm$1           \\ 
VL-TGS       & 0.40$\pm$0.49             & 7.34$\pm$1.86              & 89.2$\pm$15.1                & 25.8$\pm$10.0                  & 4.12$\pm$0.88  & 4$\pm$2    \\ 
Sem-NaVAE     & \textbf{0.90$\pm$0.30}  & 6.02$\pm$0.51  & \textbf{95.0$\pm$3.12} & 23.4$\pm$6.88    & \textbf{1.33$\pm$0.03}  & \textbf{2$\pm$1}  \\\bottomrule
\end{tabular}%
}  
\vspace{-5pt}
\end{table}

Furthermore, while NoMaD achieves a better FD by closely tracking its prior topological map, its pure visual reliance causes performance to degrade under minor lighting variations. Finally, the GPT-based method (VL-TGS) struggles primarily because its high latency prevents continuous updates based on the most recent observations. Because replanning only occurs when a trajectory is fully completed, the navigation task is inherently slower and unsuitable for real-time execution. Crucially, Sem-NaVAE's sustained high Success and reduced \#RB across these 160m-long, obstacle-dense configurations intrinsically confirms its ability to overcome unimodal limitations and systematically bypass complex topological bottlenecks (qualitatively highlighted in Fig. \ref{fig:trajectories} and Fig. \ref{fig:generated-trajectories}).

\vspace{-10pt}
\subsection{Ablation study}

To further test our system, we conducted ablation studies on the trajectory generation and selection modules, as well as on the local planner used and the update policy: (i) Two variations of NaVAE are compared, one with the standard MUSE-VAE loss \cite{musevae} (\textbf{Sem-NaVAE-BaseL}) and the other without collision loss (\textbf{Sem-NaVAE-NoCol}). (ii) As a local planner, we tested using the DWA local planner (\textbf{Sem-NaVAE-DWA}) instead of the RL based one. (iii) We tested the use of a synchronous update coupled with the generation frequency (\textbf{NaVAE-FixF}), which means the best trajectory is selected at every inference step. Two navigation experiments involving two different routes were conducted, each repeated three times. Table \ref{tab:ablation} shows the most relevant results.

\begin{table}[h]
\vspace{-5pt}
\caption{Ablation study of Sem-NaVAE.}
\label{tab:ablation}
\resizebox{\linewidth}{!}{%
\begin{tabular}{@{}ccccccc@{}}
\toprule
\textbf{Model}  & \textbf{SR $\uparrow$} & \textbf{FD $\downarrow$} & \textbf{EPT\% $\uparrow$} & \textbf{NTR\% $\downarrow$} & \textbf{$\mathbf{T_{ratio}}$ $\downarrow$} & \textbf{\#RB}$\downarrow$ \\ \midrule
Sem-NaVAE       & \textbf{6/6}             & \textbf{5.48$\pm$0.22}              & 94.5$\pm$1.36                & 23.8 $\pm$ 0.67                 & \textbf{1.34$\pm$ 0.08}                                 & \textbf{1$\pm$1}    \\
Sem-NaVAE-BaseL & 2/6             & 8.27$\pm$0.98              & 84.5$\pm$2.12                & 54.0$\pm$0.70                  & 1.56$\pm$0.21                                 & 5$\pm$2    \\
Sem-NaVAE-NoCol & 2/6             & 7.27$\pm$0.36              & 82.2$\pm$0.74                & 43.5$\pm4.94$                  & 1.41$\pm$0.05                                 & 4$\pm$1    \\
Sem-NaVAE-DWA & \textbf{6/6}             & 5.53$\pm$0.34              & \textbf{95.2$\pm$1.90}   & \textbf{23.2$\pm$2.06}                  & 1.40$\pm$0.18                               & 2$\pm$1    \\
NaVAE-FixF      & 4/6             & 5.95$\pm$0.45              & 90.7$\pm$3.5                & 23.5$\pm$1.29                  & 1.37$\pm$0.05                               & 3$\pm$2    \\ \bottomrule
\end{tabular}%
}
\vspace{-5pt}
\end{table}
Using a different loss function considerably reduces performance, primarily due to reduced prediction diversity and increased collisions.  On the other hand, using a fixed update frequency also produces worse results, as this approach may revert to a trajectory that is no better than the previous one. Given the lack of global information, this can be problematic in local minima where the robot can become trapped. Third, the performances obtained using DWA and RL planners are similar. This is as expected since the system is agnostic to the local planner, provided that it has good obstacle avoidance capabilities to account for unseen obstacles by the top level.
\vspace{-15pt}
\subsection{Qualitative results}
\begin{figure}[h]
    \centering
    \includegraphics[width=\linewidth]{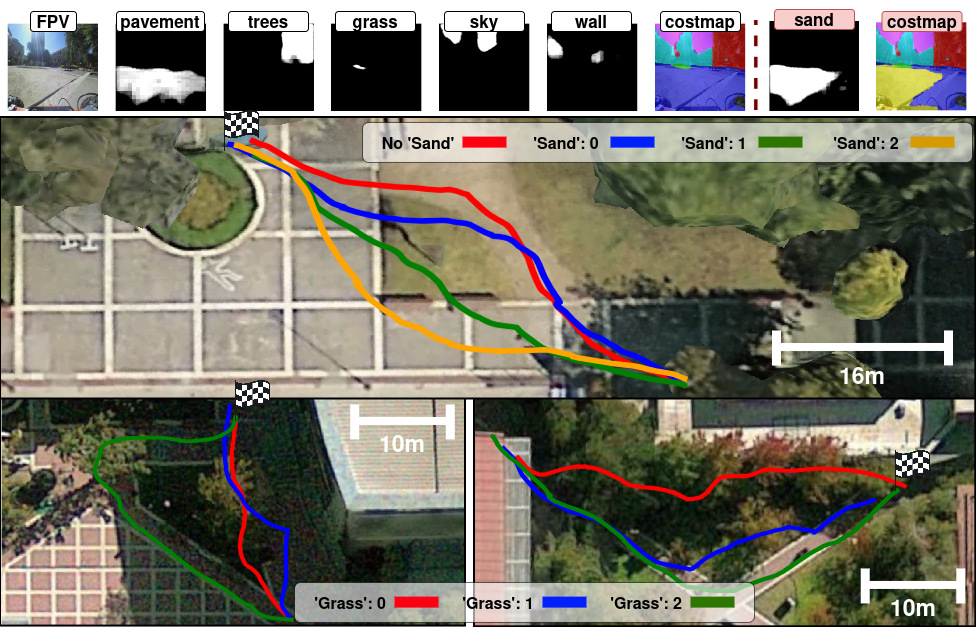}
     \vspace{-18pt}
    \caption{Comparison of executed trajectories when varying selection classes and costs.}
    \label{fig:selvar}
\end{figure}

\vspace{-2pt}
We conducted two additional experiments. First, we vary the cost of specific classes and add a new class to demonstrate the effectiveness of open-vocabulary segmentation with CLIPSeg. The bottom of Fig. \ref{fig:selvar} shows how varying the cost of the \textit{grass} class leads to different behaviors, such as taking shorter paths when the cost is low. The top of Fig. \ref{fig:selvar} shows that this behavior is also observed for a new class, such as \textit{sand}, which is considered as normal pavement when not present. When present, it can be ignored if the cost is increased.

\vspace{-10pt}

\vspace{-8pt}
Second, we conducted a long-range navigation test in an urban park using GPS odometry. Fig. \ref{fig:trajectories} shows the path followed and a close up to the selected trajectories. It can be observed that the robot can navigate successfully through a novel environment in a long range setting ($>$1km) with minimal intervention and without the need for maps, just GPS odometry. It is noticeable that the selected trajectories are not within a constant interval and that many of them are not entirely feasible; however, the selector module is capable of switching at the right moment to prevent collisions or traversing unwanted terrain.

\vspace{-8pt}
\section{Conclusion}
This work addressed the challenge of autonomous navigation in outdoor environments without pre-existing maps. By decoupling motion generation from semantic decision-making, we validated a ``Generate-and-Select'' architecture that combines the probabilistic exploration of CVAEs with the open-vocabulary segmentation capabilities of VLMs. 

Our results show that our approach successfully overcomes the limitations of unimodality in regression-based methods and the semantic inflexibility of purely geometric planners. The strict decoupling of the geometric generator from the semantic selector is the architectural property that enables both real-time operation and zero-shot adaptability, while the asynchronous update mechanism is resilient to temporal occlusions, ensuring safe operation in dynamic real-world scenarios.

Despite these promising results, we acknowledge certain limitations. First, while our real-world evaluation covers extensive routes, it is geographically limited to campus and park-like environments. Future work must validate the framework's generalization across broader environmental diversities, including varied weather, adverse lighting conditions, and denser pedestrian traffic. Second, the relatively few repeated trials per route—reflected by the variance in some quantitative metrics—suggest that further extensive testing is needed to provide a more detailed breakdown of edge-case failure modes.

\ifCLASSOPTIONcaptionsoff
  \newpage
\fi

\balance
\bibliographystyle{IEEEtran}
\bibliography{references}

@inproceedings{endtoendnav,
  title={End-to-end deep learning for autonomous navigation of mobile robot},
  author={Kim, Ye-Hoon and Jang, Jun-Ik and Yun, Sojung},
  booktitle={IEEE international conference on consumer electronics (ICCE)},
  pages={1--6},
  year={2018},
}

@inproceedings{codevilla2018endtoend,
  title={End-to-end driving via conditional imitation learning},
  author={Codevilla, Felipe and M{\"u}ller, Matthias and L{\'o}pez, Antonio and Koltun, Vladlen and Dosovitskiy, Alexey},
  booktitle={IEEE international conference on robotics and automation (ICRA)},
  pages={4693--4700},
  year={2018},
}

@article{surveyDLnav,
  title={A survey of deep learning techniques for autonomous driving},
  author={Grigorescu, Sorin and Trasnea, Bogdan and Cocias, Tiberiu and Macesanu, Gigel},
  journal={Journal of field robotics},
  volume={37},
  number={3},
  pages={362--386},
  year={2020},
  publisher={Wiley Online Library}
}

@article{predictingterrain,
  author    = {Wellhausen, Lorenz and Dosovitskiy, Alexey and Ranftl, Ren{\'e} and others},
  title     = {{Where should i walk? predicting terrain properties from images via self-supervised learning}},
  journal   = {IEEE Robotics and Automation Letters},
  volume    = {4},
  number    = {2},
  pages     = {1509--1516},
  year      = {2019},
}

@article{travmapcontext,
  author    = {Jung, Chanyoung and Shim, David Hyunchul},
  title     = {{Incorporating multi-context into the traversability map for urban autonomous driving using deep inverse reinforcement learning}},
  journal   = {IEEE Robotics and Automation Letters},
  volume    = {6},
  number    = {2},
  pages     = {1662--1669},
  year      = {2021},
}

@article{leiva2024combining,
  title={Combining RL and IL using a dynamic, performance-based
modulation over learning signals and its application to local
planning},
  author={Leiva, Francisco and Ruiz-del-Solar, Javier},
  journal={arXiv preprint arXiv:2405.09760},
  year={2024}
}

@inproceedings{hdif2023,
  title={How does it feel? self-supervised costmap learning for off-road vehicle traversability},
  author={Castro, Mateo Guaman and Triest, Samuel and Wang, Wenshan and others},
  booktitle={IEEE International Conference on Robotics and Automation (ICRA)},
  pages={931--938},
  year={2023},
}

@inproceedings{schoch2024insightinteractivenavigationsight,
  author    = {Schoch, Philipp and Yang, Fan and Ma, Yuntao and others},
  title     = {{In-sight: Interactive navigation through sight}},
  booktitle = {IEEE/RSJ International Conference on Intelligent Robots and Systems (IROS)},
  year      = {2024},
  pages     = {7794--7800},
}

@article{wvn,
  title={Fast traversability estimation for wild visual navigation},
  author={Frey, Jonas and Mattamala, Matias and Chebrolu, Nived and others},
  journal={arXiv preprint arXiv:2305.08510},
  year={2023}
}

@inproceedings{viplanner,
  author    = {Roth, Pascal and Nubert, Julian and Yang, Fan and others},
  title     = {{ViPlanner: Visual semantic imperative learning for local navigation}},
  booktitle = {IEEE International Conference on Robotics and Automation (ICRA)},
  year      = {2024},
  pages     = {5243--5249},
}

@inproceedings{chang2023goatthing,
  author    = {Chang, Matthew and Gervet, Theophile and Khanna, Mukul and others},
  title     = {{Goat: Go to any thing}}, 
  booktitle = {Robotics: Science and Systems (RSS)},
  year      = {2024},
}

@inproceedings{triest2024velociraptor,
  author    = {Triest, Samuel and Sivaprakasam, Matthew and Aich, Shubhra and others},
  title     = {{Velociraptor: Leveraging visual foundation models for label-free, risk-aware off-road navigation}},
  booktitle = {Conference on Robot Learning (CoRL)},
  year      = {2024},
}

@article{song2024tgs,
  title={Vl-tgs: Trajectory generation and selection using vision language models in mapless outdoor environments},
  author={Song, Daeun and Liang, Jing and Xiao, Xuesu and Manocha, Dinesh},
  journal={IEEE Robotics and Automation Letters},
  year={2025},
}

@article{google2024pivot,
  title={Pivot: Iterative visual prompting elicits actionable knowledge for vlms},
  author={Nasiriany, Soroush and Xia, Fei and Yu, Wenhao and others},
  journal={arXiv preprint arXiv:2402.07872},
  year={2024}
}

@inproceedings{sridhar2023nomad,
  author    = {Sridhar, Ajay and Shah, Dhruv and Glossop, Catherine and others},
  title     = {{Nomad: Goal masked diffusion policies for navigation and exploration}},
  booktitle = {IEEE International Conference on Robotics and Automation (ICRA)},
  year      = {2024},
  pages     = {12224--12230},
}

@inproceedings{gupta2018social,
  title={Social gan: Socially acceptable trajectories with generative adversarial networks},
  author={Gupta, Agrim and Johnson, Justin and Fei-Fei, Li and others},
  booktitle={IEEE conference on computer vision and pattern recognition (CVPR)},
  pages={2255--2264},
  year={2018}
}

@article{song2025vlmsocial,
  author    = {Song, Daeun and Liang, Jing and Payandeh, Amirreza and others},
  title     = {{Vlm-social-nav: Socially aware robot navigation through scoring using vision-language models}}, 
  journal   = {IEEE Robotics and Automation Letters}, 
  pages     = {508--515},
  year      = {2025},
}

@inproceedings{shah2023lmnav,
  title={Lm-nav: Robotic navigation with large pre-trained models of language, vision, and action},
  author={Shah, Dhruv and Osi{\'n}ski, B{\l}a{\.z}ej and Levine, Sergey and others},
  booktitle={Conference on robot learning (CoRL)},
  pages={492--504},
  year={2023},
}

@inproceedings{sathyamoorthy2024convoi,
  title={Convoi: Context-aware navigation using vision language models in outdoor and indoor environments},
  author={Sathyamoorthy, Adarsh Jagan and Weerakoon, Kasun and Elnoor, Mohamed and others},
  booktitle={IEEE/RSJ International Conference on Intelligent Robots and Systems (IROS)},
  pages={13837--13844},
  year={2024},
}

@inproceedings{mtg,
  author    = {Liang, Jing and Gao, Peng and Xiao, Xuesu and others},
  title     = {{MTG: Mapless trajectory generator with traversability coverage for outdoor navigation}}, 
  booktitle = {IEEE International Conference on Robotics and Automation (ICRA)}, 
  year      = {2024},
  pages     = {2396--2402},
}

@inproceedings{musevae,
  title={Muse-vae: Multi-scale vae for environment-aware long term trajectory prediction},
  author={Lee, Mihee and Sohn, Samuel S and Moon, Seonghyeon and others},
  booktitle={IEEE/CVF conference on computer vision and pattern recognition (CVPR},
  pages={2221--2230},
  year={2022}
}

@inproceedings{trajectronplusplus,
  author    = {Salzmann, Tim and Ivanovic, Boris and Chakravarty, Punarjay and Pavone, Marco},
  title     = {{Trajectron++: Dynamically-feasible trajectory forecasting with heterogeneous data}},
  booktitle = {European Conference on Computer Vision (ECCV)},
  year      = {2020},
  pages     = {683--700},
}

@INPROCEEDINGS{dtg,
  author={Liang, Jing and Payandeh, Amirreza and Song, Daeun and Xiao, Xuesu and Manocha, Dinesh},
  booktitle={IEEE/RSJ International Conference on Intelligent Robots and Systems (IROS)}, 
  title={DTG : Diffusion-based Trajectory Generation for Mapless Global Navigation}, 
  year={2024},
  volume={},
  number={},
  pages={5340-5347},
  }

@article{mosu,
  title={Mosu: Autonomous long-range robot navigation with multi-modal scene understanding},
  author={Liang, Jing and Weerakoon, Kasun and Song, Daeun and others},
  journal={arXiv preprint arXiv:2507.04686},
  year={2025}
}

@inproceedings{vlmgronav,
  author    = {Elnoor, Mohamed and Weerakoon, Kasun and Seneviratne, Gershom and others},
  title     = {{VLM-GroNav: Robot Navigation Using Physically Grounded Vision-Language Models in Outdoor Environments}}, 
  booktitle = {IEEE International Conference on Robotics and Automation (ICRA)}, 
  year      = {2025},
  pages     = {2391--2398},
}

@inproceedings{Gu2022StochasticTP,
  title={Stochastic trajectory prediction via motion indeterminacy diffusion},
  author={Gu, Tianpei and Chen, Guangyi and Li, Junlong and others},
  booktitle={IEEE/CVF conference on computer vision and pattern recognition (CVPR)},
  pages={17113--17122},
  year={2022}
}

@inproceedings{Jiang2023MotionDiffuserCM,
  author    = {Jiang, Chiyu Max and Cornman, Andre and Park, Cheol Eon and others},
  title     = {{MotionDiffuser: Controllable multi-agent motion prediction using diffusion}},
  booktitle = {IEEE/CVF Conference on Computer Vision and Pattern Recognition (CVPR)},
  year      = {2023},
  pages     = {9644--9653},
}

@inproceedings{iwae,
  author       = {Yuri Burda and
                  Roger B. Grosse and
                  Ruslan Salakhutdinov},
  editor       = {Yoshua Bengio and
                  Yann LeCun},
  title        = {Importance Weighted Autoencoders},
  booktitle    = {Conference on Learning Representations (ICLR)},
  year         = {2016},
}

@inproceedings{pointnet,
  title={Pointnet: Deep learning on point sets for 3d classification and segmentation},
  author={Qi, Charles R and Su, Hao and Mo, Kaichun and Guibas, Leonidas J},
  booktitle={IEEE conference on computer vision and pattern recognition (CVPR)},
  pages={652--660},
  year={2017}
}

@inproceedings{dino,
  title={Emerging properties in self-supervised vision transformers},
  author={Caron, Mathilde and Touvron, Hugo and Misra, Ishan and others},
  booktitle={IEEE/CVF international conference on computer vision (ICCV)},
  pages={9650--9660},
  year={2021}
}

@inproceedings{clip,
  author    = {Radford, Alec and Kim, Jong Wook and Hallacy, Chris and others},
  title     = {{Learning Transferable Visual Models From Natural Language Supervision}}, 
  booktitle = {International Conference on Machine Learning (ICML)},
  year      = {2021},
  pages     = {8748--8763},
  publisher = {PMLR}
}

@InProceedings{clipseg,
    author    = {L\"uddecke, Timo and Ecker, Alexander},
    title     = {Image Segmentation Using Text and Image Prompts},
    booktitle = {IEEE/CVF Conference on Computer Vision and Pattern Recognition (CVPR)},
    year      = {2022},
    pages     = {7086-7096}
}

@article{thrun2006stanley,
author = {Thrun, Sebastian and Montemerlo, Mike and Dahlkamp, Hendrik and Stavens, David and others},
title = {Stanley: The robot that won the DARPA Grand Challenge},
journal = {Journal of Field Robotics},
volume = {23},
number = {9},
pages = {661-692},
year = {2006}
}

@INPROCEEDINGS{Werby-hovsg, 
  author    = {Werby, Abdelrhman and Huang, Chenguang and Büchner, Martin and others}, 
  title     = {{Hierarchical open-vocabulary 3D scene graphs for language-grounded robot navigation}}, 
  booktitle = {Proceedings of Robotics: Science and Systems (RSS)}, 
  year      = {2024}, 
}

@article{gpt4,
  title={Gpt-4 technical report},
  author={Achiam, Josh and Adler, Steven and Agarwal, Sandhini and others},
  journal={arXiv preprint arXiv:2303.08774},
  year={2023}
}

@INPROCEEDINGS{behav2025,
  author={Weerakoon, Kasun and Elnoor, Mohamed and Seneviratne, Gershom and others},
  booktitle={IEEE International Conference on Robotics and Automation (ICRA)}, 
  title={Behav: Behavioral Rule Guided Autonomy Using VLMs for Robot Navigation in Outdoor Scenes}, 
  year={2025},
  volume={},
  number={},
  pages={7044-7051},
}

@inproceedings{florence2022implicit,
  title={Implicit behavioral cloning},
  author={Florence, Pete and Lynch, Corey and Zeng, Andy and others},
  booktitle={Conference on Robot Learning (CoRL)},
  pages={158--168},
  year={2022},
}

@article{frechet,
  title={Computing the Fr{\'e}chet distance between two polygonal curves},
  author={Alt, Helmut and Godau, Michael},
  journal={International Journal of Computational Geometry \& Applications},
  volume={5},
  number={01n02},
  pages={75--91},
  year={1995},
  publisher={World Scientific}
}

\end{document}